\documentclass[runningheads]{llncs}
\usepackage[T1]{fontenc}
\usepackage{graphicx,verbatim}
\usepackage{anyfontsize}
\usepackage{multirow}%
\usepackage{amsmath,amssymb,amsfonts}%
\usepackage{mathrsfs}%
\usepackage[title]{appendix}%
\usepackage{xcolor}%
\usepackage{textcomp}%
\usepackage{manyfoot}%
\usepackage{booktabs}%
\usepackage{algorithm}%
\usepackage{algorithmicx}%
\usepackage{algpseudocode}%
\usepackage{listings}%
\begin{document}
\title{Towards Robust Algorithms for Surgical Phase Recognition via Digital Twin Representation}
\author{Hao Ding\inst{1}\and
Yuqian Zhang\inst{1} \and
Wenzheng Cheng\inst{1} \and
Xinyu Wang\inst{1} \and
Xu Lian\inst{1} \and
Chenhao Yu\inst{1} \and
Hongchao Shu\inst{1} \and
Ji Woong Kim\inst{1} \and
Axel Krieger\inst{1} \and \\
Mathias Unberath\inst{1}}
\authorrunning{H. Ding et al.}
\authorrunning{H. Ding et al.}
\institute{Johns Hopkins University, Baltimore MD 21218, USA 
\email{hding15@jhu.edu,unberath@jhu.edu} \\}

% \author{Anonymized Authors}  %% Added for anonymized MICCAI 2025 submission
% \authorrunning{Anonymized Author et al.}
% \institute{Anonymized Affiliations \\
%     \email{email@anonymized.com}}

\maketitle              % typeset the header of the contribution
\begin{abstract}
Surgical phase recognition (SPR) is an integral component of surgical data science, enabling high-level surgical analysis. End-to-end trained neural networks that predict surgical phase directly from videos have shown excellent performance on benchmarks. However, these models struggle with robustness due to non-causal associations in the training set. Our goal is to improve model robustness to variations in the surgical videos by leveraging the digital twin (DT) paradigm -- an intermediary layer to separate high-level analysis (SPR) from low-level processing. As a proof of concept, we present a DT representation-based framework for SPR from videos. The framework employs vision foundation models with reliable low-level scene understanding to craft DT representation. We embed the DT representation in place of raw video inputs in the state-of-the-art SPR model. The framework is trained on the Cholec80 dataset and evaluated on out-of-distribution (OOD) and corrupted test samples. Contrary to the vulnerability of the baseline model, our framework demonstrates strong robustness on both OOD and corrupted samples, with a video-level accuracy of 80.3 on a highly corrupted Cholec80 test set, 67.9 on the challenging CRCD dataset, and 99.8 on an internal robotic surgery dataset, outperforming the baseline by 3.9, 16.8, and 90.9 respectively. We also find that using DT representation as an augmentation to the raw input can significantly improve model robustness. Our findings lend support to the thesis that DT representations are effective in enhancing model robustness. Future work will seek to improve the feature informativeness and incorporate interpretability for a more comprehensive framework.

% Authors must provide keywords and are not allowed to remove this Keyword section.

\end{abstract}

\section{Introduction}
\begin{figure}[t]
\centering{
    \includegraphics[width=0.95\textwidth]{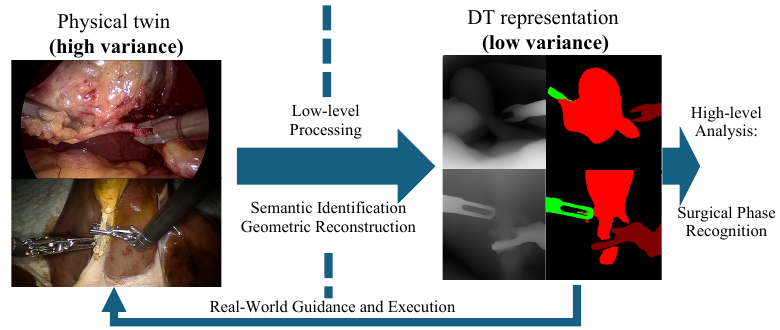}
    \caption{Illustration of the DT paradigm. DT Paradigm demonstrates a clear seperation between low-level processing with high-level analysis based on DT representation}\label{fig:dt}
}
\end{figure}
Surgical phase recognition (SPR) is a pivotal task in surgical data science, providing essential insights for surgical workflow analysis, skill assessment, and numerous other applications. With the rise of deep learning, end-to-end feed-forward networks~\cite{TwinandaSMMMP17endonet,JinD0YQFH18sv-rcnet,YiJ19hfom,GaoJLDH21transsvnet,liu2023skit,yang2024surgformer} have demonstrated strong performance on SPR benchmarks~\cite{TwinandaSMMMP17endonet,wang2022autolaparo}. Despite excellent performance, the end-to-end learning fashion for feed-forward networks (FFNs) poses the challenge of model robustness~\cite{drenkow2021systematic} when facing domain gap or non-adversarial corruptions~\cite{ding2024segstrong}. The lack of reliability of the proposed algorithms hindered the clinical translation of research achievements for surgical data science~\cite{maier2022surgical,ding2024digital}. Researchers started to take low-level vision tasks such as instrument segmentation and assess the models' robustness by adding corruptions to the input image~\cite{colleoni2020synthetic}. Ding et al. added non-adversarial physical corruptions to the test data~\cite{ding2024segstrong,ding2022carts,ding2023rethinking} taking advantage of the precise replayability of the da Vinci Research Kit~\cite{kazanzides2014open}. However, to our best knowledge, there is a lack of exploration of the model's robustness for high-level analysis of surgical procedures like surgical phase recognition.

The digital twin (DT) paradigm, as illustrated in Figure~\ref{fig:dt}, constructs and maintains computational representations of real-world environments. It serves as an intermediary layer between the video pixels and SPR, achieving a separation between high-level analysis and low-level processing. From the real-world observations (physical twin) with high variance, DT applies semantic identification and geometric reconstruction techniques such as segmentation, depth estimation, 3D reconstruction, and pose estimation to extract a digital counterpart, a DT representation~\cite{ding2024digital}. High-level analysis model is then trained and conducted on this low variance representation to reduce non-causal learning when training feed-forward networks, thereby enhancing the model’s robustness. Nowadays, DT has received increasing attention, and extracting DT representations has become more feasible due to the emergence of vision foundation models for low-level processing, like the segment anything model 2 (SAM2)\cite{ravi2024sam} and DepthAnything\cite{yang2024depth}. These models have shown impressive zero-shot generalization and robustness~\cite{shen2024performance}, thanks to their large model capacity, vast training datasets, and advanced self- or semi-supervised learning mechanisms. Previous works~\cite{shu2023twin,hein2024creating,killeen2024stand,kleinbeck2024neural,ding2024towards} focus on the extraction of the DT representation from external tracking devices and markers or visual input. Ding et al.~\cite{ding2024towards} applied the DT representation in an embodied surgical system and improved the robustness of the surgical automation, providing evidence of the effectiveness of the DT representation in improving the robustness of the downstream application.

In this work, we introduce a DT representation–based framework to improve robustness in SPR tasks. We harness the strong performance and robustness of vision foundation models by utilizing SAM2~\cite{ravi2024sam} and DepthAnything~\cite{yang2024depth} to extract basic DT representations. These structured representations then replace video inputs in Surgformer~\cite{yang2024surgformer} model, enabling SPR from surgical videos via the DT with enhanced robustness. We train our framework based on the Cholec80~\cite{TwinandaSMMMP17endonet} dataset and evaluate its performance and robustness on both original and corrupted images from the Cholec80 test dataset as well as two out-of-distribution (OOD) sets: CRCD~\cite{koh2024crcd} and an internal cholecystectomy dataset for robotics training acquired using the da Vinci Research Kit (dVRK) on ex vivo porcine specimens. Our results show that our framework demonstrates strong robustness against image corruptions and domain shifts where the end-to-end trained baseline models fail. We also successfully augmented the baseline model with the DT representation, achieving significantly improved robustness. These findings support the effectiveness of using DT representations to enhance model robustness, paving the way for more reliable high-level analysis for surgical data and accelerating the clinical translation of research outcomes. The experiment design and corresponding materials like a 10-class instance segmentation of the full Cholec80 dataset will also contribute to the following research.

Our contribution can be summarized as follows: (1) A robust SPR framework from surgical video via a DT representation; (2) An approach to form and apply a DT representation to state-of-the-art models; and (3) A comprehensive study of the robustness of the proposed framework on corrupted and OOD test data as well as the experimental materials for future research. 
\section{Methods}
\begin{figure}[t]
\centering{
    \includegraphics[width=1\textwidth]{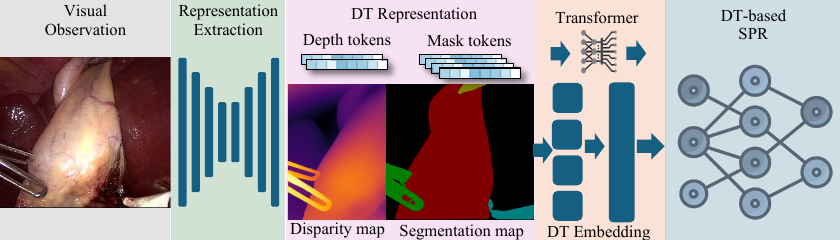}
    \caption{Illustration of the surgical phase
recognition framework via DT representation.}\label{fig:method}
}
\end{figure}

Figure~\ref{fig:method} presents an overview of our proposed robust surgical phase recognition framework using DT representations. The framework comprises three primary modules: representation extraction, DT embedding, and DT-based SPR architecture - DT Former. In the representation extraction module, raw visual inputs, such as surgical videos, are processed using vision foundation models to extract explicit fundamental representations including depth maps, segmentation masks, and their corresponding latent space tokens. These representations are further processed in the DT embedding module to form DT tokens. The final module, DT Former, applies the existing SPR model (Surgformer), taking DT tokens for training and inference. All modules will be detailed in the following subsections.

\subsection{Representation Extraction}
For representation extraction, we employ SAM2~\cite{ravi2024sam} for instance segmentation and DepthAnything~\cite{yang2024depth} for monocular depth estimation.

\noindent\textbf{Instance segmentation.}
We define 10 classes, as outlined in Table~\ref{tab:instances}. These include one target tissue (gallbladder), one specimen bag, and six types of surgical instruments: grasper, bipolar, hook, scissors, clipper, and irrigator. To distinguish between multiple graspers that appear simultaneously, we differentiate them based on their insertion positions.
\begin{table}[t]
\caption{Instance list for segmentation. Each pair is the instance name and id.}\label{tab:instances}%
\resizebox{\columnwidth}{!}{
\begin{tabular}{cc|cc|cc|cc|cc}
\toprule%
gallbladder & 1 & left grasper & 2 & top grasper & 3 & right grasper & 4 & bipolar & 5 \\
\midrule
hook & 6 & scissors & 7 & clipper & 8 & irrigator & 9 & specimen bag & 10\\
\bottomrule
\end{tabular}
}
\end{table}
For instance segmentation, we use SAM2~\cite{ravi2024sam}, providing a point prompt for each instance at the frame where it first appears. SAM2 utilizes a Vision Transformer (ViT)~\cite{DosovitskiyB0WZ21vit} as the image encoder and incorporates a memory attention module to integrate spatial and temporal information from previous and prompted frames. The point prompt is encoded using a prompt encoder, and a mask decoder processes the encoded prompt and image features to generate mask predictions while updating memory. In our framework, we take both the generated mask and corresponding mask tokens output from SAM2's mask decoder and retain them for the following steps. The mask tokens for each frame are represented as 10 object entries where each entry is a vector of length 257, representing one instance from Table~\ref{tab:instances}. The first 256 dimensions of the vector are the output mask token from SAM2 where the last digit represent the existence of the object in the frame. 

\noindent\textbf{Depth estimation.} We apply the DepthAnything~\cite{yang2024depth} model to each individual frame for depth estimation. The model follows an encoder-decoder architecture, estimating relative disparity from monocular images. The predicted disparity is normalized to a 0-1 scale, which we use to represent relative depth. The output tokens of the encoder and decoder are further processed by averaging spatial dimensions, producing the depth token used in subsequent steps.

\subsection{DT Embedding}

For each frame, the raw representation consists of 10 binary segmentation masks for each class and a normalized disparity map. We format the segmentation masks into a 10-channel, one-hot encoded tensor, then concatenate the disparity map as the 11th channel. A 2D convolution layer is then applied to the sequence of frames, transforming the tensor into a spatial-temporal token sequence. To incorporate depth token and mask token, we applied two transformers for further processing. For the mask token, the first 256 dimensions of each 257-dimensional entry are projected into an embedding space, and a sinusoidal positional embedding is added to encode their positions. The embedded mask tokens are then appended by a learnable token and fed into self-attention blocks with the last digit of then entry as the mask for attention layers. The depth tokens undergo a similar embedding process, except without the binary mask for attention.

\subsection{DT Former}

We applie Surgformer~\cite{yang2024surgformer} as the prediction model to determine the surgical phase for each frame. Surgformer processes a video clip of fixed length, appending a classification embedding to aggregate features for phase prediction. It integrates hierarchical temporal attention (HTA) into the TimesFormer~\cite{bertasius2021space} encoder to capture both global and local temporal information. The processed classification embedding from the final layer is passed to a linear head, predicting the phase of the last frame in online prediction settings. In our architecture, namely DT Former, we replace tokens from patch embedding of the raw video clip with our DT tokens. The mask token is early fused with the cls token via element-wise addition prior to the attention layers. By contrast, the depth token is late fused with the cls token via element-wise addition after the final layers. 
\section{Experiment}
In this section, we present the experiments conducted to assess the robustness of our proposed method. We conduct two primary experiments: model robustness and ablation studies.
 %robustness against image corruptions
For both experiments, our method and the baseline model are trained on the public Cholec80 dataset~\cite{TwinandaSMMMP17endonet}. In the model robustness experiment, we evaluate our method on the original  Cholec80 test set and corrupted Cholec80-C test set with combined corruption (Hue transformation, brightness, contrast), as well as the CRCD~\cite{koh2024crcd} and an internal robotics training dataset, both serving as OOD test sets. 
\subsection{Experiment Setting}

\noindent\textbf{Dataset.}
We train our method and baselines on the Cholec80 dataset~\cite{TwinandaSMMMP17endonet}, which contains 80 cholecystectomy videos, each annotated with 7 surgical phases. The dataset is evenly split into training/validation and testing sets. 
The CRCD~\cite{koh2024crcd} dataset includes videos recorded during ex vivo pseudo-cholecystectomy procedures on pig livers from which we take 5 videos performing gallbladder dissection for evaluation. The internal robotics training dataset consists of 10 videos capturing robotic ex-vivo cholecystectomy training on the dVRK platform, specifically performing cutting and clipping tasks. 

%The corrupted samples can be found 
\noindent\textbf{Evaluation Metrics.}
In this work, we evaluate performance using four key metrics: video-level accuracy, phase-level precision, recall, and Jaccard index. Video-level accuracy represents the mean percentage of correctly recognized frames for each video. Given the class imbalance, phase-level metrics offer further insights. Precision is defined as the ratio of true positives to total predictions, recall is the ratio of true positives to actual instances, and the Jaccard index measures the overlap between predictions and ground truths. For each phase, we collect the total number of true positives, false positives, and false negatives, and compute precision, recall, and Jaccard. Finally, we calculate the mean for these metrics across all phases. For the CRCD~\cite{koh2024crcd} and robotics training test data, since there record specific procedure belonging to one specific phase for each dataset, we only report recall for phase-level metrics.

% \subsubsection{Baseline Models}
\noindent\textbf{Model Setting.}
We apply the ViT-Tiny encoder for both the SAM2~\cite{ravi2024sam} model and depthAnything~\cite{yang2024depth} model and use the officially provided pre-trained model for inference. For the Surgformer~\cite{yang2024surgformer} model, we adopt the official online SPR implementation with an input frame sequence length of 16, a sampling rate of 4, and a learning rate of 0.0005. We train each model for 30 epochs with a batch size of 8. The AutoAugmentation, random erasing, and random spatial sampling are also applied to the training of the baseline model.
\begin{figure}[t]
\centering{
    \includegraphics[width=\textwidth]{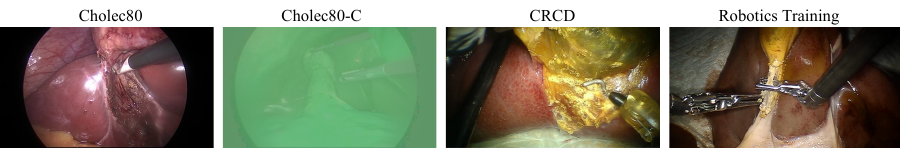}
    \caption{Visual examples of the original and corrupted Cholec80~\cite{TwinandaSMMMP17endonet}, CRCD~\cite{koh2024crcd},and the robotics training dataset.}\label{fig:dataset}
}
\end{figure}
\begin{table}[t]
\caption{Robustness experiment results. Cholec80-C denotes the corrupted Cholec80.  RT denotes robotics training datasets."D" and "M" means depth-only and mask-only. }\label{tab:experiment}
\begin{tabular*}{\textwidth}{@{\extracolsep\fill}lcccccccccccc}
\toprule%

\bf{OOD} & \multicolumn{4}{c}{Cholec80~\cite{TwinandaSMMMP17endonet}} & \multicolumn{4}{c}{Cholec80-C~\cite{TwinandaSMMMP17endonet}} 
& \multicolumn{2}{c}{CRCD~\cite{koh2024crcd}} 
& \multicolumn{2}{c}{RT} 
\\ 
\cmidrule{2-5}
\cmidrule{6-9}
\cmidrule{10-11}%
\cmidrule{12-13}
Methods 
& Acc & Prec & Rec & Jac 
& Acc & Prec & Rec & Jac 
& Acc  & Rec
& Acc  & Rec
\\
\midrule
Surgformer
& 92.2 & 87.1 & 87.6 & 77.8 
& 76.4 & 70.4 & 78.4 & 57.2 
& 56.0 & 56.6 & 8.9 & 8.8
\\
DT Former - D
& 83.4 & 75.3 & 77.7 & 62.3
& 37.3 & 50.3 & 40.9 & 20.6 
& 38.4 & 36.1 & 99.1 & 99.1
\\
DT Former - M
& 75.2 & 72.4 & 72.5 & 56.4 
& 74.6 & 71.5 & 72.1 & 55.6 
& 51.1 & 44.3 & 91.1 & 90.8
\\
DT Former
& 85.5 & 79.9 & 78.2 & 67.2 
& \bf{80.3} & \bf{76.3} & \bf{76.0} & \bf{60.7}
& \bf{67.9} & \bf{64.1} & \bf{99.8} & \bf{99.8}
\\
\bottomrule
\end{tabular*}
\end{table}
\subsection{Model Robustness}

\noindent\textbf{OOD Generalization.} As shown in Figure~\ref{fig:dataset}, although these datasets depict the same procedure, there is a notable appearance gap between the Cholec80~\cite{TwinandaSMMMP17endonet}, CRCD~\cite{koh2024crcd}, and robotic cholecystectomy datasets, making them well-suited for OOD evaluation. The results, presented in the OOD section of Table~\ref{tab:experiment}, highlight that while the Surgformer baseline performs strongly on the Cholec80 test data, it struggles significantly on both OOD test sets, especially on the Robotics Training dataset where it failed to identify cutting and clipping phases. By contrast, our model demonstrates robust performance, attaining $67.9$ and $99.8$ video-level accuracy, respectively on the CRCD~\cite{koh2024crcd} and robotics training datasets, outperforming Surgformer by a large margin. These results underscores our model's robustness to OOD variations.

\noindent\textbf{Robustness against Corruption.} We apply combined corruption from hue transformation, brightness adjustment, and contrast alteration to the first 10 test videos to form a corrupted test set. The corruptions are selected and implemented according to the ImageNet-C practice~\cite{hendrycks2019benchmarking}. Examples of these corrupted images are shown in Figure~\ref{fig:dataset}. As seen in the corruption results in Table~\ref{tab:experiment}, the baseline model’s performance substantially deteriorates under the corruption. In contrast, our framework demonstrates strong robustness overall, with only a slight drop in performance, We attribute this decline primarily to inaccuracies in segmentation mask and depth prediction caused by the corruptions.

\begin{table}[t]
\caption{Results of baseline model with DT augmentation.}\label{tab:augmentation}
\begin{tabular*}{\textwidth}{@{\extracolsep\fill}lcccccccc}
\toprule%

& \multicolumn{4}{c}{Cholec80-C~\cite{TwinandaSMMMP17endonet}} 
& \multicolumn{2}{c}{CRCD~\cite{koh2024crcd}} 
& \multicolumn{2}{c}{RT} 
\\ 
\cmidrule{2-5}
\cmidrule{6-7}
\cmidrule{8-9}
Methods 
& Acc & Prec & Rec & Jac 
& Acc  & Rec
& Acc  & Rec
\\
\midrule
Surgformer
& 76.4 & 70.4 & 78.4 & 57.2 
& 56.0 & 56.6 & 8.9 & 8.8
\\
Surgformer with DT-Aug
& 79.4 & 76.4 & 78.6 & 61.0
& 68.3 & 65.3 & 99.9 & 99.9
\\
\bottomrule
\end{tabular*}
\end{table}

\subsection{Ablation Study}
For the ablation study, we first explore the contribution of depth and segmentation information in the DT Former to assess the contribution of depth and segmentation. Then we explore the possibility of using DT representation as an augmentation to raw video input.

\noindent\textbf{Components.} We create depth-only and segmentation-only models by removing one modality at a time from the input. The results, presented in Table~\ref{tab:experiment}, shows that, between the two variants, taking depth information solely as input generates better benchmark performance but solely taking mask information as input generates better robustness. We suppose this is because depth representation contains more fine-grained details, making the learning process easier while the mask information is more abstract thus the performance of the corresponding foundation models is more consistent when facing corruptions. Combining mask and depth makes a good complementary for each other, improving comprehensive model performance.

\noindent\textbf{DT representation augmented baseline.} DT representation-based model demonstrates strong robustness in the previous experiment. In this study, we explore another possibility of incorporating the DT representation as an augmentation into the raw input to improve the baseline model's robustness. We concatenate our 11-channel spatial DT input with the raw rgb input and adapt the depth and history tokens into the baseline model to form the augmented baseline. The results, presented in Table~\ref{tab:augmentation} show that the DT augmentation improves the robustness of the baseline model on both corrupted and OOD test samples. This further validates the idea of applying DT representation for better model robustness in high-level analysis tasks.  
\section{Discussion}
Our DT-based SPR framework demonstrates strong robustness in high-level surgical analysis tasks, underscoring the efficacy of the DT paradigm as an intermediary layer for enhancing surgical data science research. The advent of vision foundation models provides a highly generalizable and reliable low-level processing pipeline, facilitating the extraction and formation of DT representations. This capability holds promise for fostering more generalizable and interpretable systems in surgical data science, potentially accelerating the clinical translation of these research advancements. As a proof of concept and an initial exploration of DT-based high-level analysis, further development is required to create a more comprehensive and practical framework for SPR and other high-level analytical tasks. On one hand, despite using the same model capacity, on clean data, the benchmark performance of our framework remain below the  state-of-the-art methods, indicating that it may not yet extract sufficient information for SPR. Thus, Future research should investigate additional low-level features, such as scene flow, which can provide informative data for SPR and be reliably extracted. On the other hand, while this work focuses on model robustness, the interpretability offered by the DT representation has not yet been fully explored. Such interpretability could be highly relevant to clinical settings by clarifying decision-making processes and instilling greater trust in automated systems, and therefore is a direction to further explore in future efforts.

\section{Conclusion}
We propose a surgical video analysis framework using a DT representation. This framework demonstrates strong robustness against OOD and corrupted test samples, validating the effectiveness of high-level surgical analysis via DT representation and the reliability of low-level representation extraction through vision foundation models. Our work highlights a promising research direction for developing more robust algorithms in surgical data science, moving beyond recent end-to-end deep learning approaches. To create a more comprehensive framework, future research should focus on enhancing the informativeness of DT-based feature representations and improving the efficacy of the feature extraction pipeline. Additionally, incorporating explainable AI techniques to enhance interpretability is likely beneficial for facilitating clinical translation.  
\begin{credits}
\subsubsection{\ackname}
This research is supported by a collaborative research agreement with the MultiScale Medical Robotics Center at The Chinese University of Hong Kong. 
\end{credits}
\bibliographystyle{splncs04}
\bibliography{main}
\end{document}